\documentclass[11pt,a4paper]{article}
\usepackage[hyperref]{emnlp2020}

\usepackage{times}
\usepackage{latexsym}
\usepackage{xcolor}
\usepackage{soul}
\usepackage{microtype}
\usepackage{graphicx}
\usepackage{subfigure}
\usepackage{booktabs} 
\usepackage{multirow}
\usepackage{url}
\usepackage{amsmath}
\usepackage{float}
\usepackage{booktabs}
\usepackage{amssymb}
\usepackage{mathtools, nccmath}
\usepackage{bm}
\usepackage{capt-of}
\usepackage{cleveref}
\usepackage{algorithm}
\usepackage{algorithmic}
\usepackage{url}
\usepackage{amsthm}
\definecolor{darkgreen}{rgb}{0.0, 0.5, 0.1}
\mathchardef\mhyphen="2D
\DeclareMathOperator*{\concat}{\scalebox{1}[1.5]{$\parallel$}}

\aclfinalcopy 
\def\aclpaperid{1396} 

\title{A Knowledge-Driven Approach to Classifying Object and \\Attribute Coreferences in Opinion Mining}

\author{Jiahua Chen$^{\dag}$,~ Shuai Wang$^{\ddag}$\thanks{~~Work done while at University of Illinois at Chicago}~,~ Sahisnu Mazumder$^{\dag}$,~ Bing Liu$^{\dag}$\\
  $^{\dag}$Department of Computer Science, University of Illinois at Chicago, USA \\
  $^{\ddag}$Amazon AI\\
  \texttt{jiahuaqy@gmail.com,~shuaiwanghk@gmail.com}\\
  \texttt{sahisnumazumder@gmail.com,~liub@uic.edu} \\}

\date{}

\begin{document}
\maketitle

\begin{abstract}
\label{sec:abstract}
Classifying and resolving coreferences of objects (e.g., \textit{product names}) and attributes (e.g., product \textit{aspects}) in opinionated reviews 
is crucial for improving the opinion mining performance. However, the task is challenging as one often needs to consider domain-specific knowledge (e.g., \textit{iPad is a tablet and has aspect resolution}) to identify coreferences in opinionated reviews. Also, compiling a hand-crafted and curated domain-specific knowledge base for each domain is very time consuming and arduous. This paper proposes an approach to \textit{automatically} mine and leverage domain-specific knowledge for classifying objects and attribute coreferences. The approach extracts domain-specific knowledge from \textit{unlabeled} review data and trains a knowledge-aware neural coreference classification model to leverage (useful) domain knowledge together with general commonsense knowledge for the task. Experimental evaluation on real-world datasets involving five domains (product types) shows the effectiveness of the approach.
\end{abstract}

\section{Introduction}
\label{sec:introduction}
Coreference resolution (CR) aims to determine whether two mentions (linguistic referring expressions) corefer or not, i.e., they refer to the same entity in the discourse model 
\citep{jurafsky2000speech,ding2010resolving,atkinson2015improving,lee2017end,lee2018higher,joshi2019bert,zhang2019knowledge}. The set of coreferring expressions forms a coreference chain or a cluster. Let's have an example:

\begin{quote}
    \textbf{[S1]}~ \textit{{\color{blue}I} bought {\color{red} a green Moonbeam} for {\color{blue}myself}}. ~
    \textbf{[S2]}~ \textit{{\color{blue}I} like \underline{{\color{red} its} voice} because \underline{it} is loud and long}.
\end{quote}

Here all colored and/or underlined phrases are mentions. Considering S1 (sentence-1) and S2 (sentence-2), the three mentions ``\textit{I}", ``\textit{myself}" in S1 and ``\textit{I}" in S2 all refer to the same person and form a cluster. Similarly, ``\textit{its}" in S2 refers to the object ``\textit{a green Moonbeam}" in S1 and the cluster is \{``\textit{its}"~(S2), ``\textit{a green Moonbeam}"~(S1) \}. The mentions ``\textit{its voice}" and ``\textit{it}" in S2 refer to the same attribute of the object ``\textit{a green Moonbeam}" in S1 and form cluster \{``\textit{its voice}"~(S2), ``\textit{it}"~(S2)\}.

CR is beneficial for improving 
many down-stream NLP tasks such as question answering~\citep{dasigi2019quoref}, dialog systems~\citep{quan2019gecor}, entity linking~\citep{DBLP:conf/acl/KunduSFH18}, and opinion mining~\citep{nicolov2008sentiment}. Particularly, in opinion mining tasks~\cite{liu2012sentiment,wang2016attention,zhang2018deep,ma2020entity},~\citet{nicolov2008sentiment} reported performance improves by 10\% when CR is used. The study by \citet{ding2010resolving} also supports this finding. Considering the aforementioned example, without resolving ``\textit{it}" in S2, it is difficult to infer the opinion about the attribute ``\textit{voice}" (i.e., the \textit{voice}, which ``\textit{it}" refers to, is ``\textit{loud and long}"). Although CR plays such a crucial role in opinion mining, only limited research has been done for CR on opinionated reviews. CR in opinionated reviews (e.g.,  Amazon product reviews) \textit{mainly} concerns about resolving coreferences involving objects and their attributes. 
The objects in reviews are usually the names of products or services while attributes are aspects of those objects~\citep{liu2012sentiment}.

Resolving coreferences in text broadly involves performing three tasks  (although they are often performed jointly or via end-to-end learning): (1) identifying the list of mentions in the text (known as \textbf{\textit{mention detection}}); (2) given a pair of candidate mentions in text, making a binary classification decision: \textit{coreferring} or \textit{not} (referred to as \textbf{\textit{coreference classification}}), and (3) grouping coreferring mentions (referring to the same discourse entity) to form a coreference chain (known as \textbf{\textit{clustering}}). In reviews, \textbf{mention detection} is equivalent to extracting entities and aspects in reviews which has been widely studied in opinion mining or sentiment analysis~\cite{hu2004mining,qiu2011opinion,DBLP:conf/naacl/XuLSY19,DBLP:conf/acl/LuoLLZ19,wang2018disentangling,DBLP:journals/ipm/DragoniFR19,DBLP:journals/cluster/AsgharKZAK19}. Also, once the coreferring mentions are detected via classification, clustering them could be straightforward\footnote{Given a text (context), if pairs ($m$, $p$), ($m$, $q$) are classified as co-referring mentions, then $m$, $p$, $q$ belong to same cluster.}. Thus, following~\citep{ding2010resolving}, \textit{we only focus on solving the \textbf{coreference classification} task in this work, which we refer to as the \textbf{object and attribute coreference classification} (OAC2) task onwards.} We formulate the OAC2 problem as follows.

\textbf{Problem Statement.} Given a review \textbf{\textit{text}} $u$ (context), an \textbf{\textit{anaphor}}\footnote{The term anaphor used in this work does not have to be the same as defined in other related studies, as here it can also appear before $m$ though rarely. We still name it as anaphor for simplicity, mainly following~\citep{ding2010resolving}.}
$p$ and a \textbf{\textit{mention}} $m$ which refers to either an object or an attribute (including their position information), our goal is to \textit{predict whether the anaphor $p$ refers to mention $m$, denoted by a binary class $y \in \{0, 1\}$}. Note. an anaphor here can be a \textit{pronoun} (e.g.,  ``it") or \textit{definite noun phrase} (e.g.,  ``the clock") or \textit{ordinal} (e.g.,  ``the green one"). 

In general, to classify coreferences, one needs  intensive knowledge support. For example, to determine that ``\textit{it}" refers to ``\textit{its voice}" in S2, we need to know that ``\textit{voice}" can be described as ``\textit{loud and long}" and ``\textit{it}" can not refer to ``\textit{a green Moonbeam}" in S1, since ``\textit{Moonbeam}" is a clock 
which cannot be described as ``\textit{long}". 

Product reviews contain a great many such domain-specific concepts like brands (e.g.,  ``\textit{Apple}" in the laptop domain), product name (e.g.,  ``\textit{T490}" in the computer domain), and aspects (e.g.``\textit{hand}" in the alarm clock domain) that often do not exist in general knowledge bases (KBs) like WordNet \citep{miller1998wordnet}, ConceptNet~\citep{singh2002open}, etc. Moreover, even if a concept exists in a general KB, its semantics may be different than that in a given product domain. For example, ``\textit{Moonbeam}" in a general KB is understood as ``\textit{the light of the moon}" or \textit{the name of a song}, rather than a clock (in the alarm clock domain). To encode such domain-specific concepts, 
we need to mine and feed domain knowledge (e.g.,  ``\textit{clock}'' for ``\textit{Moonbeam}'', ``\textit{laptop}'' for ``\textit{T490}'') to a coreference classification model. Existing CR methods \citep{zhang2019knowledge} do not leverage such domain knowledge and thus, often fail to resolve such co-references that require explicit reasoning over domain facts. 

In this paper, we propose to \textit{automatically} mine such domain-specific knowledge from \textit{unlabeled} reviews and 
leverage the useful pieces of the extracted domain knowledge together with the (general/comensense) knowledge from general KBs to solve the OAC2 task\footnote{The unlabeled data are from the same source as the annotated data (i.e., the same domain, but without labels), which can ensure the reliability of the domain knowledge as well as the coverage of mention words. With the domain-specific knowledge
mined, the meaning of a mention in a certain domain can be better understood (by a model) with the support of its relevant mentions (extracted from the self-mined KB).}. Note the extracted domain knowledge and the general knowledge from the existing general KBs are both considered as candidate knowledge. To leverage such knowledge, we design a novel knowledge-aware neural coreference classification model that selects the useful (candidate) knowledge with attention mechanism. We discuss our approach in details in Section 3.

The main contributions of this work can be summarized:
\begin{enumerate}
    \vspace{-0.2cm}
    \item We propose a knowledge-driven approach to solving OAC2 in opinionated reviews. Unlike existing approaches that mostly dealt with general CR corpus and pronoun resolution, we show the importance of leveraging domain-specific knowledge for OAC2.
    \vspace{-0.2cm}
    \item We propose a method to automatically mine  domain-specific  knowledge and design a novel knowledge-aware coreference classification model that leverages both domain-specific and general knowledge.
    \vspace{-0.2cm}
    \item We collect a new review dataset\footnote{\url{https://github.com/jeffchen2018/review_coref}} with five domains or product types (including both unlabeled and labeled data) for evaluation. Experimental results show the effectiveness of our approach. 
\end{enumerate}

\section{Related Work}
\label{sec:relatedwork}
Coreference resolution has been a long-studied problem in NLP. Early approaches were mainly rule-based~\citep{hobbs1978resolving} and feature-based~\citep{ding2010resolving,atkinson2015improving} where researchers focused on leveraging lexical, grammatical properties and semantic information. Recently, 
end-to-end solutions with deep neural models \citep{lee2017end,lee2018higher,joshi2019bert} have dominated the coreference resolution research. But they did not use external knowledge. 

Conisdering CR approaches that use external knowledge, ~\citet{aralikatte2019rewarding}
solved CR task by incorporate knowledge or information in reinforcement learning models. ~\citet{emami2018generalized}  solved the  binary choice coreference-resolution task by leveraging information retrieval results from search engines.
~\citet{zhang2019incorporating,zhang2019knowledge} solved pronoun coreference resolutions by leveraging contextual, linguistic features, and external knowledge where knowledge attention was utilized. However, these works did not deal with opinionated reviews and also did not mine or use domain-driven knowledge.

In regard to CR in opinion mining, ~\citet{ding2010resolving} formally introduced the OAC2 task for opinionated reviews, which is perhaps the only prior study on this problem. However, it only focused on classifying coreferences in comparative sentences (not on all review sentences). We compare our approach with \citep{ding2010resolving} in Section 4. 

Many existing general-purpose CR datasets are not suitable for our task, which include MUC-6 and MUC-7 \citep{hirschman1998appendix}, ACE~\citep{doddington2004automatic}, OntoNotes \citep{pradhan2012conll}, and  WikiCoref \citep{ghaddar2016wikicoref}. \citet{bailey2015winograd} proposed an alternative Turing test, comprising a binary choice CR task that requires significant commonsense knowledge. ~\citet{yu2019you} proposed visual pronoun coreference resolution in dialogues that require the model to incorporate image information. These datasets are also not suitable for us as they are not opinionated reviews. We do not focus on solving pronoun resolution here because, for opinion text such as reviews, discussions and blogs, personal pronouns mostly refer to one person~\citep{ding2010resolving}. Also, we aim to leverage domain-specific knowledge on (unlabeled) domain-specific reviews to help the CR task which has not been studied by any of these existing CR works. 

\section{Proposed Approach}
\label{sec:model}

\begin{figure*}
\centering
\includegraphics[width=6.25in]{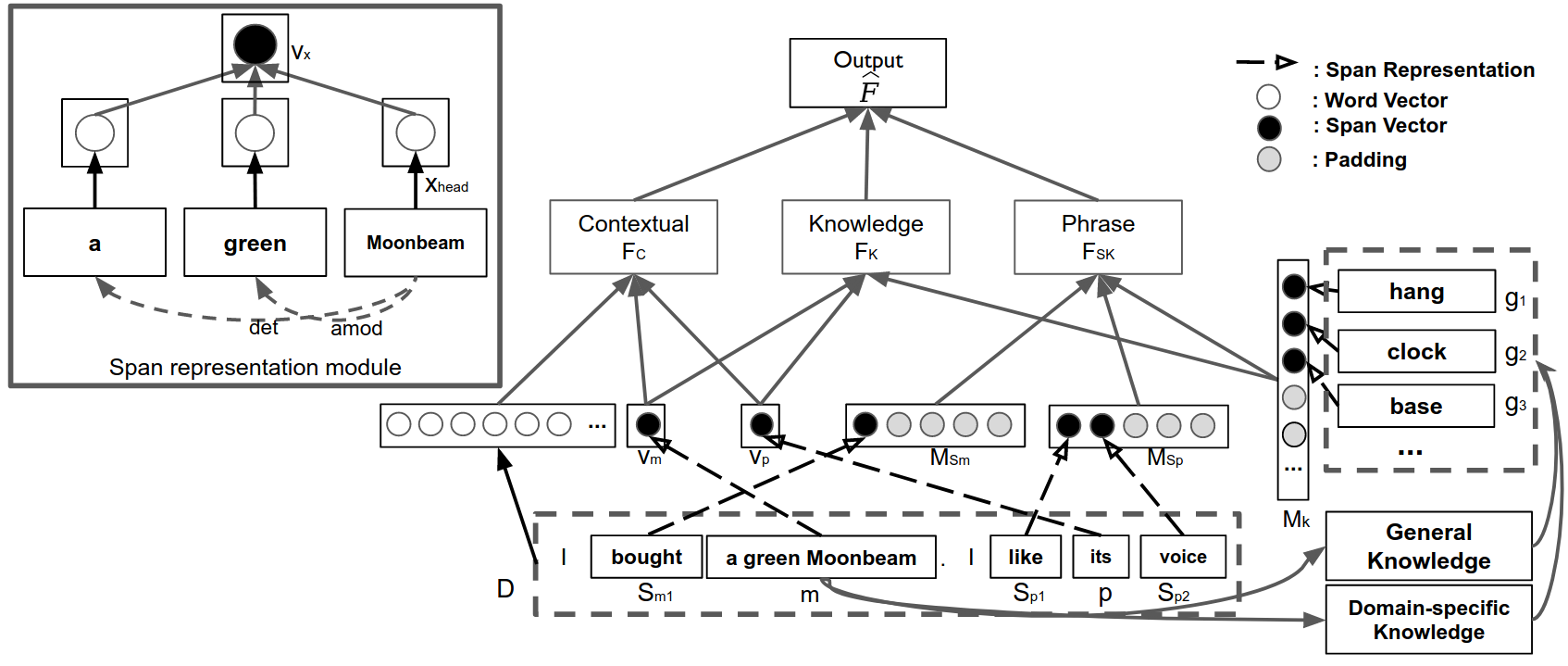}
\caption{\label{fig:model}The architecture of our knowledge-driven OAC2 model. 
}
\vspace{-0.2cm}
\end{figure*}

\textbf{Model Overview.} Our approach consists of the following three main steps: \textbf{(1) knowledge aquisition}, where given the (input) pair of mention $m$ (e.g.,  ``\textit{a green Moonbeam}'') and anaphor $p$ (e.g., ``\textit{it}'') and the context $t$ (i.e., the review text), we acquire candidate knowledge involving $m$, denoted as $K_m$.~$K_m$ consists of both domain knowledge (mined from unlabeled reviews) as well as general knowledge (compiled from existing general KBs) (discussed in Section 3.1). 
Next, in \textbf{(2) syntax-based span representation}, we extract syntax-related phrases for mention $m$ and anaphor $p$. Syntax-related phrases are basically noun phrases, verbs or adjectives that have a dependency relation\footnote{
We use \url{spacy.io} for dependency parsing, POS tagging and Named Entity Recognition (NER) in our implementation.} with $m$ (or $p$). For example,  ``\textit{bought}" is a syntax-related phrase of the mention ``\textit{a green Moonbeam}'' and ``\textit{like}" and ``\textit{voice}" are two syntax-related phrases for the anaphor ``it" in the example review text in Section 1. Once the syntax-related phrases are extracted and the candidate knowledge is prepared for $m$ and $p$, we learn vector representations of the phrases and the knowledge (discussed in Section 3.2), which are used in step-3.
Finally, in \textbf{(3) knowledge-driven OAC2 model}, we select and leverage useful candidate domain 
knowledge together with general knowledge to solve the OAC2 task. Figure~\ref{fig:model} shows our model architecture. Table~\ref{table:notation} summarizes a (non-exhaustive) list of notations, used repeatedly in subsequent sections.

\begin{table}
\centering
\caption{\small Summary of notations (non-exhaustive list)}
\label{table:notation}
\scalebox{0.84}{
\begin{tabular}{lp{6.5cm}}
\hline
$d$ & a domain\\
$t$ & a review text or context\\
$m$ & a mention \\
$p$ & an anaphor \\
$K_m$ & (domain+general) knowledge involving $m$ for domain $d$ \\
$K^d_m$ & domain knowledge involving $m$ for $d$\\
$S_m$ & syntax-related phrases of $m$ \\
$S_p$ & syntax-related phrases of $p$\\
$T_d$ & labeled reviews in $d$\\
$\overline{T}_d$ & unlabeled reviews in $d$\\
\hline
\end{tabular}
}
\label{tab:plain}
\vspace{-0.3cm}
\end{table}

\subsection{Knowledge Acquisition}
\label{subsec:extraction}
~~~\textbf{Domain Knowledge Mining.}
Given the mention $m$, we first split the mention into words. Here, we only keep the words that satisfy one of the following two conditions\footnote{When only using two features of words, we already achieve good results. More features are left for future work.}: (1) a word is a noun (determined by its POS tag); (2) a word is part of a named entity (by NER). For example,  ``\textit{a westclox clock}" will result in words ``\textit{westclox}" and ``\textit{clock}". We use the mention words as the keys to search a domain knowledge base (KB) to retrieve domain knowledge for the mention $m$. 

To construct the domain KB, we use unlabeled review data in the particular domain. Specifically, all unlabeled sentences that contain mention words are extracted. Next, we collect domain knowledge for $m$ as $K^d_m$, where $K^d_m$ = $\{k^d_{m,1}, k^d_{m,2}, ... \}$. The elements in $K^d_m$ are phrases of nouns, adjectives, and verbs co-occurring with $m$ in the unlabeled review sentences.  

\textbf{Domain Knowledge Filtering.}
Some domain knowledge (i.e., co-occurring phrases) can be too general to help reason over the mention. For example,  given mention ``\textit{Moonbeam}", the verb ``\textit{like}" can be related to any objects or attributes and thus, is not a very useful knowledge for describing the mention. To filter such unimportant phrases from $K_m^d$, we use $tf\mhyphen idf$~\citep{aizawa2003information} scoring. 

Given mention $m$ and a phrase $k \in K_m^d$, we compute $tf\mhyphen idf$ score of $k$, denoted as $tf\mhyphen idf_{k}$ as given below: 
\vspace{-1mm}
\begin{align}
    tf_{k} &= \frac{C_{k}}{max_{k'\in K^d_{m}} C_{k'}}\\
    idf_{k} &= \log \frac{\vert \overline{T}_d \vert} {\vert \lbrace t' \in \overline{T}_d : k \in t' \rbrace \vert}
\end{align}
\vspace{-0.35cm}
\begin{align}
    tf\mhyphen idf_{k} &= tf_{k} \cdot  idf_{k}
\end{align}
where $C_{k}$ denotes the co-occurrence count of phrase $k$ with $m$ in unlabeled domain reviews $\overline{T}_d$ and $\vert{\cdot}\vert$ denotes set count.  We retain phrase $k$ in $k_m^d$, if $tf\mhyphen idf_k \geq \rho$, where $\rho$ is a (empirically set) threshold value. 

\textbf{General Knowledge Aquisition.} General Knowledge bases like ConceptNet, WordNet, etc. store facts as triples of the form  ($e_1$, $r$, $e_2$), denoting entity $e_1$ is related to entity $e_2$ by a relation $r$. e.g., (``\textit{clock}", ``\textit{UsedFor}", ``\textit{set an alarm}"). 

To acquire and use general knowledge for mention $m$, we first split $m$ into words (in the same way as we do during domain knowledge construction) and use these words as keywords to retrieve triples such that one of the entities (in a given triple) contains a word of $m$. Finally, we collect the set of entities (from the retrieved triples) as general knowledge for $m$, by selecting the other entity (i.e., instead of the entity involving a mention word) from each of those retrieved triples.


\subsection{Syntax-based Span Representation}
\label{subsec:span}
Once the domain-specific and general knowledge for mention $m$ is acquired, we extract all syntax-related phrases for $m$ and anaphor $p$ from review text $t$ (see ``Model Overview" in Section 3). We denote the syntax-related phrases of $m$ and $p$ as $S_m$ and $S_p$ respectively. 

We represent mention, anaphor, the syntax-related phrases,
and also the phrases of knowledge from domain-specific and general KBs as \textbf{spans} (a continuous sequence of words), and learn a vector representation for each span (we call it a \textbf{span vector}) based on the embeddings of words
that compose the span. The span vectors are then used by our knowledge-driven OAC2 model (discussed in Section 3.3) for solving the OAC2 task. Below, we discuss the span vector representation learning for a given span (corresponding to a syntax-related phrase or a phrase in KB).

We use BERT~\citep{devlin2018bert} to learn the vector representation for each span. To encode the words in a span, we use BERT's WordPiece tokenizer. Given a span $x$, let $\{x_i\}_{i=1}^{N_1}$ be the output token embeddings of $x$ from BERT, where $N_1$ is the total number of word-piece tokens for span $x$. 

BERT is a neural model consisting of stacked attention layers. To incorporate the syntax-based information, we want the head of a span and words that  have a modifier relation to the head to have higher attention weights. To achieve the goal, we adopt syntax-based attention \citep{he2018effective}. The weight of a word in a span depends on the dependency parsing result of the span. Note, the dependency parsing of a span is different from what is described in Section~\ref{subsec:extraction}. The dependency parsing in Section~\ref{subsec:extraction} extracts the relation between chunks of words while here we extract relations between single words. 

An example has been shown in top left corner of Figure~\ref{fig:model}. The head of ``\textit{a green Moonbeam}" is ``\textit{Moonbeam}" that we want to have the highest attention weight when computing the embedding of the span. The distance of (``\textit{a}", ``\textit{Moonbeam}") and (``\textit{green}", ``\textit{Moonbeam}") considering the dependency path are both 1. 

To learn the span vector $v_x$ for span $x$, we first compute the attention weights $b_i$'s for each $x_i$, as:
\begin{align}
    f_i &= FFN_1([x_i, x_{head}, x_i\odot x_{head}])\\
    a_i &= \begin{cases} 
    \frac{1}{2^{l_i}}\cdot exp(f_i),  &if ~l_i\le L\\
    0, & otherwise\\
    \end{cases}\\
    b_i &= \frac{a_i}{\sum_{j=1}^{N_1} a_j}
\end{align}
where $FFN_1$ is a feed-forward layer that projects the input into a score $f_i$, $\odot$ is element-wise multiplication, $[,]$ is concatenation, $x_{head}$ is the head of the span, $l_i$ is the distance to the head along the dependency path, $L$ is the attention window size. 

Next, we learn the attention-based representation of the span $x$, denoted as $\hat{x}$ as:
\begin{align}
    \hat{x} &= \sum_{i=1}^{N_1} b_i \cdot x_i
\end{align}

Finally, we concatenate the start and end word embeddings of the span $x_{start}$ and $x_{end}$, attention-based representation $\hat{x}$ and a length feature $\phi(x)$ following~\citep{lee2017end} to learn span vector $v_x$: 
\begin{align}
    v_x &= FFN_2([x_{start}, x_{end}, \hat{x}, \phi(x)]).
\end{align}
where $FFN_2$ is a feed-forward layer.

\subsection{Knowledge-driven OAC2 Model}
\label{subsec:model}
The knowledge-driven OAC2 model leverages the syntax-related phrases together with the domain knowledge and general knowledge to solve the OAC2 task. The model first computes three relevance scores: \textbf{(a)} a contextual relevance score $F_C$ between $m$ and $p$, \textbf{(b)} a knowledge-based relevance score $F_K$ between $m$ and $p$, and \textbf{(c)} a relevance score $F_{SK}$ between knowledge and syntax-related phrases (see Figure~\ref{fig:model}) and then, these scores are summed up to compute the final prediction score $\hat{F}$, as shown below:
\begin{equation}\label{2-def}
\scalebox{1.0}{$
\hat{F}=~ sigmoid(F_C+F_K+F_{SK})$}
\end{equation}
 
\vspace{1mm}
\textbf{(a) Contextual Relevance Score ($F_C$).} $F_C$ is computed based on the context $t$, mention $m$ and anaphor $p$. We use BERT to encode $t$. Let the output BERT embeddings of words in $t$ be $\{t_i\}_{i=1}^{N_2}$, where $N_2$ is length of $t$. Also, let the span vector representations of $m$ and $p$ are $v_m$ and $v_p$ respectively. Then, for each $v \in \{v_m, v_p\}$, we compute cross attention between $t$ and $v$ as follows:
\begin{align}
    g_i &= FFN_3([t_i, v, t_i\odot v])\\
    w_{i}^v &= \frac{e^{g_i}}{\sum_{j=1}^{N_2} e^{g_j}}\cdot t_i
\end{align}
where $FFN_3$ is a feed-forward layer.

We learn the interaction of $\{t_i\}_{i=1}^{N_2}$ with $v_m$ and $v_p$ to get attention-based vector representations $\{w_i^m\}_{i=1}^{N_2}$ and $\{w_i^p\}_{i=1}^{N_2}$ for $m$ and $p$ respectively. Next, we concatenate these vectors and their point-wise multiplication for each context word, sum up the concatenated representations and feed it to a feed-forward layer to compute $F_C \in \mathcal{R}^{1 \times 1}$:
\begin{align}
F_C &= FFN_4(\sum_{i=1}^{N_2} [w_i^{m}, w_i^{p}, w_i^{m}\odot w_i^{p}])
\end{align}
where $FFN_4$ is a feed-forward layer.

\vspace{1mm}
\textbf{(b) Knowledge-based Relevance Score ($F_K$).} The OAC2 model leverages the external knowledge to compute a relevance score $F_K$ between $m$ and $p$. Let $v_m$ and $v_p$ be the span vectors for $m$ and $p$ and  $\{v_{i}^K\}_{i=1}^{N_3}$ be the span vectors for phrases in $K_m$ (see Sec 3.1 and Table 1), where $N_3$ is size of $K_m$. Then, we compute $F_K$ using $v_m$, $v_p$ and $\{v_{i}^K\}_{i=1}^{N_3}$ as discussed below. 

To leverage external knowledge information, we first learn cross attention between the mention and the knowledge as:
\begin{align}
    h_i &= FFN_5([v_{i}^K, v_m, v_{i}^K\odot v_m])\\
    c_i &= \frac{e^{h_i}}{\sum_{j=1}^{N_3} e^{h_j}}
\end{align}
where $FFN_5$ is a feed-forward layer.

Next, we learn an attention-based representation $\hat{v}_m$ of mention $m$ as:
\begin{align}
    \hat{v}_m &= \sum_{i=1}^{N_3} c_i \cdot v_{i}^K
\end{align}
We now concatenate $v_m$, $v_p$, the attention-based representation $\hat{v}_m$ and learn interaction between them to compute $F_K \in \mathcal{R}^{1 \times 1}$ as:

\begin{align}
\scalebox{0.99}{$
F_K = FFN_6([v_m, v_p, \hat{v}_m, v_p \odot \hat{v}_m, v_p \odot \hat{v}_m])$
}
\end{align}
where $FFN_6$ is a feed-forward layer.

\vspace{1mm}
\textbf{(c) Syntax-related Phrase Relevance Score ($F_{SK}$).}
$F_{SK}$ measures the relevance between the knowledge (i.e., phrases) in $K_m$ and the syntax-related phrases in $S_m$ ($S_p$) corresponding to $m$ ($p$).

Let $v_i^{K}$ be the span vector for $i^{th}$ phrase in $K_m$ and $v_i^{m}$ ($v_i^{p}$) be the span vector for $i^{th}$ phrase in $S_m$ ($S_p$). Then, we concatenate these span vectors row-wise to form matrices $M_{K}$ = $v_i^{K}\concat_{i=1}^{N_3}$ $\in \mathcal{R}^{N_3 \times d}$, $M_{Sm}$ = $v_i^{m}\concat_{i=1}^{N_4} \in \mathcal{R}^{N_4 \times d}$ and $M_{Sp}$ = $v_i^{p}\concat_{i=1}^{N_5} \in \mathcal{R}^{N_5 \times d}$ respectively, where $\concat_{i=1}^{Q}$ denotes concatenation of $Q$ elements, $d$ is dimension of span vector, $N_4$ ($N_5$) is size of $S_m$ ($S_p$). 

Next, we learn interaction between these matrices using scaled dot attention~\citep{vaswani2017attention} as:
\begin{align}
    \tilde{M}_{Sm} &= softmax(\frac{M_{Sm}M_{K}^T}{\sqrt{d}})M_{K}\\
    \tilde{M}_{Sp} &= softmax(\frac{M_{Sp}M_{K}^T}{\sqrt{d}})M_{K}
\end{align}

Finally, the syntax-related phrase relevance score $F_{SK} \in \mathcal{R}^{1 \times 1}$ is computed as:
\begin{align}
    F_{SK} =FFN_8( FFN_7(\tilde{M}_{Sm}\tilde{M}_{Sp}^T))
\end{align}
where $FFN_7$ and $FFN_8$ are two feed-forward network layers. 


\vspace{1mm}
\textbf{Loss Function.}  As shown in Equation~\ref{2-def}, given  three scores $F_C$, $F_K$, and $F_{SG}$, we sum them up and then feed the sum into a sigmoid function to get the final prediction $\hat{F}$. The proposed model is trained in an end-to-end manner by minimizing the following cross-entropy loss $\mathcal{L}$:
\begin{equation}\label{1-def}
\mathcal{L} = -\frac{1}{N}\sum_i^N[y_i \cdot \log(\hat{F_i}) + (1-y_i)\cdot \log(1-\hat{F_i})]
\end{equation}
where, $N$ is the number of training examples and $y_i$ is the ground truth label of $i^{th}$ training example.

\begin{table}
\small
\centering
\caption{\small Dataset Statistics. \#R means the number of annotated reviews and \#E indicates total entities that refer to objects or attributes. P and N stand for positive and negative examples and the values under them are the numbers of those examples.}
\label{table:dataset}
\begin{tabular}{lcccccccc}
\hline
  \multirow{2}{*}{\textbf{Domain}} &   \multirow{2}{*}{\textbf{\#R}} &  \multirow{2}{*}{\textbf{\#E}} &   \multicolumn{2}{c}{\textbf{Train}}  & \multicolumn{2}{c}{\textbf{Dev}} & \multicolumn{2}{c}{\textbf{Test} }\\ 
  \cline{4-9}
& & & \textbf{P}  & \textbf{N} &  \textbf{P}  & \textbf{N} &  \textbf{P}  & \textbf{N}\\
\hline
alarm    & 100  & 924  &	647	&	1533	&	96	&	243	&	89	&	187  \\
camera   & 100  & 871 &	632	&	1709	&	69	&	160	&	83	&	174    \\
cellphone& 100  & 938  &	679	&	1693	&	62	&	148	&	73	&	189  \\
computer & 100  & 1035 &	703	&	1847	&	86	&	227	&	112	&	273   \\
laptop   & 100  & 893  &	641	&	1618	&	88	&	244	&	77	&	209  \\
\hline
\end{tabular}
\label{tab:plain}
\vspace{-0.35cm}
\end{table}

\section{Experiments}
\label{sec:experiment}
We evaluate our proposed approach using five datasets 
associated with five different domains: (1) \textit{alarm clock}, (2) \textit{camera}, (3) \textit{cellphone}, (4) \textit{computer}, and (5) \textit{laptop} and perform both quantitative and qualitative analysis in terms of predictive performance and domain-specific knowledge usage ability of the proposed model.

\subsection{Evaluation Setup}
\label{subsec:setup}
\vspace{1mm}
{\large \textbf{Labelled Data Collection.}}  We use the product review dataset\footnote{\url{https://www.cs.uic.edu/~zchen/downloads/ICML2014-Chen-Dataset.zip}} from ~\citet{chen2014topic}, where each product (domain) has 1,000 unlabeled reviews. For each domain, we randomly sample 100 reviews, extract a list of (\textit{mention}, \textit{anaphor}) pairs from each of those reviews and label them manually with ground truths. That is, given a review text and a candidate (\textit{mention}, \textit{anaphor}) pair, we assign a binary label to denote whether they co-refer or not. In other words, we view each labeled example as a triple ($u, m, p$), consisting of the \textbf{context} $u$, a \textbf{mention} $m$ and an \textbf{anaphor} $p$. Considering the review example (in Section 1), the triple (``\textit{I bought $\dots$ loud and long}", ``\textit{a green Moonbeam}", ``\textit{its}") is a positive example, since "\textit{a green Moonbeam}" and "\textit{its}" refers to the \textbf{same entity} (i.e., they are in the same coreference cluster). Negative examples are naturally constructed by selecting $m$ and $p$ from two different clusters under the same context like (``\textit{I bought $\dots$ loud and long}", ``\textit{a green Moonbeam}", ``\textit{its voice}"). 

Next, we randomly split the set of all labeled examples (for a given domain) into 80\% for training, 10\% as development, and rest 10\% as test data. The remaining 900 unlabeled reviews form the \textit{unlabeled domain corpus} is used for domain-specific knowledge extraction (as discussed in Section 3.1). All sentences in reviews and (\textit{mention}, \textit{anaphor}) pairs were annotated by two annotators independently who strictly followed the MUC-7 annotation standard~\citep{hirschman1998appendix}. The Cohen's kappa coefficient between two annotators is 0.906. When disagreement happens, two annotators adjudicate to make a final decision. 
Table~\ref{table:dataset} provides the statistics of labeled dataset used for training, development and test for each of the five domains. 

\vspace{1mm}
\noindent
{\large \textbf{Knowledge Resources.}} We used three types of knowledge resources as listed below. The first two are general KBs, while the third one is our mined domain-specific KB. 

\textbf{1. Commonsense knowledge graph (OMCS)}. We use the open mind common sense (OMCS) KB as general knowledge~\citep{singh2002open}. OMCS contains 600K crowd-sourced commonsense triplets such as (\textit{clock}, \textit{UsedFor}, \textit{keeping time}). We follow \citep{zhang2019knowledge} to select highly-confident triplets and build the OMCS KG consisting of total 62,730 triplets.

\textbf{2. Senticnet}~\citep{cambria2016senticnet}. Senticnet is another commonsense knowledge base that contains 50k concepts associated with affective properties including sentiment information. To make the knowledge base fit for deep neural models, we concatenate SenticNet embeddings with BERT embeddings to extend the embedding information.
     
\textbf{3. Domain-specific KB}. This is mined from the unlabeled review dataset as discussed in Sec~\ref{subsec:extraction}. 

\vspace{1mm}
\noindent
{\large \textbf{Hyper-parameter Settings.}} Following the previous work of~\citep{joshi2019bert,lee2018higher}, we use (Base) BERT\footnote{{\color{red}\url{https://storage.googleapis.com/bert_models/2020_02_20/uncased_L-12_H-768_A-12.zip}}} embeddings of context and knowledge representation (as discussed in Section 3). The number of training epochs is empirically set as 20. We train five models on five datasets separately, because the domain knowledge learned from a certain domain may conflict with that from others. 
Without loss of generality and model extensibility, we use the same set of hyper-parameter settings for all models built on each of the five different domains. We select the best model setting based on its performance on the development set, by averaging five F1-scores on the five datasets. The best model uses maximum length of a sequence
as 256, dropout as 0.1, learning rate as $3e^{-5}$ with linear decay as $1e^{-4}$ for parameter learning, and $\rho = 5.0$ (threshold for \textit{tf-idf})  in domain-specific knowledge extraction (Section 3.1). The tuning of the other baseline models is the same as we do for our model.

\vspace{1mm}
\noindent
{\large \textbf{Baselines.}} We compare following state-of-the art models from existing works on CR task:

\textbf{(1) Review CR}~\citep{ding2010resolving}: A review-specific CR model that incorporates opinion mining based features and linguistic features. 

\textbf{(2) Review CR+BERT}: For a fairer comparison, we further combine BERT with features from~\citep{ding2010resolving} as additional features. Specifically, we combine the context-based BERT to compute $F_C(m,p)$ (see Section 3.3 (a)).

\textbf{(3) C2f-Coref}~\citep{lee2018higher}: A state-of-the-art end-to-end model that leverages contextual information and pre-trained Glove embeddings.

\textbf{(4) C2f-Coref+BERT}~\citep{joshi2019bert}: This model integrates BERT into C2f-Coref. We use its   $independent$ setting which uses non-overlapping segments of a paragraph, as it is the best performing model in ~\citet{joshi2019bert}.  
 
\textbf{(5) Knowledge+BERT}~\citep{zhang2019knowledge}: This is a state-of-the-art knowledge-base model, which leverages different types of general knowledge and contextual information by incorporating an attention module over knowledge. General knowledge includes the aforementioned OMCS, linguistic feature and selectional preference knowledge extracted from Wikipedia. To have a fair comparison, we replace the entire LSTM-base encoder with BERT-base transformer.
    
To accommodate the aforementioned baseline models into our settings, which takes \textit{context}, \textit{anaphor}, and \textit{mention} as input and perform binary classification, we change the input and output of the baseline models, i.e., the models compute a score between mention and anaphor  and feeds the score to a sigmoid function to get a score within $[0, 1]$. Note, this setting is consistently used for all candidate models (including our proposed model). 

\vspace{1mm}
\noindent
{\large \textbf{Evaluation Metrics.}} As we aim to solve the OAC2 problem, a focused coreference classification task, we use the standard evaluation metrics \textit{F1-score} $(F1)$, following the same setting of the prior study~\cite{ding2010resolving}. In particular, we report positive (+ve) F1-score [F1(+)]. 
The average +ve F1-score is computed over five domains. 

\begin{table}
\small
\centering
\caption{\label{table:result}\small Performance (+ve F1 scores) of all models on all test datasets. Here, ``cam'', ``com'', ``lap'' are the abbreviation for ``camera'', ``computer'', ``laptop'' respectively.}
\label{tab:plain}
\begin{tabular}{lcccccc}
\hline
\textbf{Model}&\textbf{alarm}&\textbf{cam}&\textbf{phone}&\textbf{com}&\textbf{lap}&\textbf{average}\\
\hline
Review CR& 58.2&60.5&57.7&59.6&58.9&58.98\\ \hline
Review CR& \multirow{2}{*}{67.2} & \multirow{2}{*}{69.3} & \multirow{2}{*}{67.0}& \multirow{2}{*}{68.4}& \multirow{2}{*}{66.7}& \multirow{2}{*}{67.72}\\
+BERT &&&&&&\\ \hline
C2f-Coref&68.8& 70.1& 67.2& 69.5& 67.4&68.60\\ \hline
C2f-Coref& \multirow{2}{*}{70.2} & \multirow{2}{*}{71.6} & \multirow{2}{*}{68.6}& \multirow{2}{*}{71.3}& \multirow{2}{*}{68.2}& \multirow{2}{*}{69.98}\\
+BERT &&&&&&\\ \hline
Knowledge& \multirow{2}{*}{72.0} & \multirow{2}{*}{73.4} & \multirow{2}{*}{71.8}& \multirow{2}{*}{72.6}& \multirow{2}{*}{70.0}& \multirow{2}{*}{71.96}\\
+BERT &&&&&&\\ \hline
Our model   
&\textbf{73.6}&\textbf{74.5}&\textbf{72.4}&\textbf{73.8}&\textbf{71.3}&\textbf{73.12}\\
\hline
\end{tabular}
\vspace{-0.35cm}
\end{table}

\subsection{Results and Analysis}
\label{subsec:analysis}
\vspace{1mm}
\noindent
{\large \textbf{Comparison with baselines.}} Table~\ref{table:result} reports F1 scores of all models for each of  five domains and average F1 over all domains. We observe the following: (1) Overall, our model performs the best considering all five domains, outperforming the no-knowledge baseline model C2f-Coref+BERT by 3.14\%. On the  cellphone domain, our model outperforms it by 3.8\%. (2) Knowledge+BERT turns out to be the strongest baseline, outperforming the other three baselines, which also shows the importance of leveraging external knowledge for the OAC2 task. However, our model achieves superior performance over Knowledge+BERT which indicates leveraging domain-specific knowledge indeed helps. (3) C2f-Coref+BERT achieves better scores than C2f-Coref and Review CR. This demonstrates that both representation (using pre-trained BERT) and neural architectures are important for feature fusions in this task.    

\begin{table}
\small
\centering
\caption{\label{table:ablation}\small Performance of our model with different types of knowledge or module removed (-). $\Delta$ F1(+) is the performance difference between our model and model with module remove.} 
\begin{tabular}{p{1.7cm}p{2.9cm}cc}
\hline
\textbf{Comparison}&\textbf{Model} & \textbf{Avg. F1(+)} & \textbf{$\Delta$ F1(+)}\\\hline
& Our model      &73.12 & 0.00\\\hline
Knowledge&-OMCS knowledge         &72.28& 0.84\\
source&-Domain knowledge       &72.22& 0.90\\
&-Senticnet              &72.82&0.30\\
&-all knowledge          &70.56&2.56\\\hline
Score&-context $F_{c}$          &71.14&1.98\\
&-knowledge $F_K$         &71.80&1.48\\
&-phrase $F_{SG}$ &72.58&0.56\\\hline
attention&-syntax-based attention &72.50&0.62\\
&\quad+dot attention          &72.96&0.16\\
\hline
\end{tabular}
\vspace{-0.35cm}
\end{table}

\vspace{2mm}
\noindent
{\large \textbf{Ablation study.}} To gain further insight, we ablate various components of our model with the results reported in Table~\ref{table:ablation}. For simplicity, we only show the average F1-scores on the five domain datasets. The results indicate how each knowledge resource or module contributes, from which we have the following observations.

\begin{enumerate}
   \vspace{-0.2cm}
    \item From comparison Knowledge resources in Table~\ref{table:ablation}, we see that domain knowledge contributes the most. General OMCS knowledge also contributes 0.84 to the model on average, so general knowledge is still needed. Senticnet contributes the least as it is more about sentiment rather than the relatedness between mentions. If we remove all knowledge sources (-all knowledge), performance drop becomes the highest which shows the importance of leveraging external knowledge in OAC2. 
    \vspace{-0.2cm}
    \item Considering comparisons of various types of scores in Table~\ref{table:ablation}, we see that the disabling the use of context score $F_C$ has the highest drop in performance, showing the importance of contextual information for this task. Disabling the use of knowledge scores $F_{G}$ and $F_{SG}$ also impact the predictive performance of the model, by causing a drop in performance.
    \vspace{-0.2cm}
    \item From the comparison of attention mechanism for span representation in Table~\ref{table:ablation}, we see that, before summing up the embedding of each word of the span, the attention layer is necessary. Note, we use the selected attention instead of popular dot attention in~\citep{vaswani2017attention} during span representation. The influence of the syntax-based attention layer is slightly  better than the dot attention layer. Therefore, we use the selected attention for better interpretability.
    \vspace{-0.2cm}
\end{enumerate}

\begin{table}
\small
\centering
\caption{\label{table:casestudy}\small A test example from alarm domain with class probability distributions by three models during prediction.}
\begin{tabular}{p{2.77cm}p{4.5cm}} \hline
\textbf{Context} & ...after I bought {\color{blue} (a green Moonbeam} for myself ... potential buyer also should know that , as with (the other Westclox clock), {\color{blue} (the clock)} also have (a gold band) ... \\ \hline
\textbf{(Mention, Anaphor)}& (\textit{a darkgreen Moonbeam}, \textit{the clock}) \\\hline
\textbf{Domain knowledge}& drop, {\color{darkgreen}hang}, {\color{darkgreen}clock}, {\color{darkgreen}put}, {\color{darkgreen}alarm}, clear, beautiful, expensive, worthwhile ...\\ \hline
\textbf{Our model} & (0: 0.47, 1: 0.53)\\ \hline
\textbf{Knowledge+BERT} & (0: 0.87, 1: 0.13)\\ \hline
\textbf{C2f-coref+BERT} & (0: 0.79, 1: 0.21)\\ \hline
\end{tabular}
\end{table}

\begin{table}
\small
\centering
\caption{\label{table:tfidf}\small An example showing the domain knowledge extraction quality of our model from laptop domain.}
\begin{tabular}{p{3.33cm}p{4cm}} \hline
\textbf{Mention (Domain)}& ~windows (laptop)\\\hline
\textbf{Extracted knowledge (before filtering)} & ~keep, like, product, battery, fast, {\color{darkgreen}microsoft}, {\color{darkgreen} system}, {\color{darkgreen} upgrade}, {\color{darkgreen}xp}, laptop.. \\\hline
\textbf{Candidate knowledge (after filtering by $tf\mhyphen idf$)} & ~{\color{darkgreen} microsoft}, {\color{darkgreen} system}, {\color{darkgreen} upgrade}, {\color{darkgreen}xp}, laptop..\\\hline
\end{tabular}
\vspace{-0.35cm}
\end{table}

\vspace{1mm}
\noindent
{\large \textbf{Qualitative Evaluation.}} We first give a real example to show the effectiveness of our model by comparing it with two baseline models C2f-coref$+$BERT and Knowledge$+$BERT. Table~\ref{table:casestudy} shows a sample in the \textit{alarm} domain. Here the major difficulty is to identify ``Moonbeam" as a ``clock". Knowledge$+$BERT fails due to its lack of domain-specific knowledge. C2f-coref$+$BERT fails as well because it simply tries to infer from contextual information only, where there is no domain knowledge support. In contrast, with our domain-specific knowledge base incorporated, ``Moonbeam" can be matched to the knowledge like ``clock", ``alarm", and ``hang" which are marked with green color. So our model successfully addresses this case. In other words, in our model, not only the mention ``a green Moonbeam" but also syntax-related phrase ``a gold band" of ``the clock" will be jointly considered in reasoning. We can see the modeling superiority of our knowledge-aware solution. Table~\ref{table:tfidf} shows the effectiveness of our extraction module introduced in Section~\ref{subsec:extraction}, especially the usage of $tf\mhyphen idf$ to filter out useless knowledge. 

\section{Conclusion}
\label{sec:conclusion}
This paper proposed a knowledge-driven approach for object and attribute coreference classification in opinion mining. The approach can automatically extract domain-specific knowledge from unlabeled data and leverage it together with the general knowledge for solving the problem. We also created a set of annotated opinionated review data (including 5 domains) for object and attribute coreference evaluation. Experimental results show that our approach achieves state-of-the-art performance. 

\section*{Acknowledgments}
This work was supported in part by two grants from National Science Foundation: IIS-1910424 and IIS-1838770, and one research gift from Tencent.

\bibliographystyle{acl_natbib}
\bibliography{emnlp2020}

\end{document}